# Performance Evaluation of Road Traffic Control Using a Fuzzy Cellular Model

Bartłomiej Płaczek

Faculty of Transport, Silesian University of Technology,
Krasińskiego 8, 40-019 Katowice, Poland
bartlomiej.placzek@polsl.pl

**Abstract.** In this paper a method is proposed for performance evaluation of road traffic control systems. The method is designed to be implemented in an on-line simulation environment, which enables optimisation of adaptive traffic control strategies. Performance measures are computed using a fuzzy cellular traffic model, formulated as a hybrid system combining cellular automata and fuzzy calculus. Experimental results show that the introduced method allows the performance to be evaluated using imprecise traffic measurements. Moreover, the fuzzy definitions of performance measures are convenient for uncertainty determination in traffic control decisions.

## 1 Introduction

On-line simulation is an effective approach for performance evaluation and optimisation of adaptive road traffic control [5]. The term on-line means that the simulation is synchronised with real time and it is adjusted to traffic data. In this technique real-time traffic measurements are the main inputs to a traffic model, which is used to predict future evolution of the traffic flow. Applying a predictive traffic model, the performance of traffic control is estimated in terms of travel times, delays, queue lengths, numbers of stops and others. These performance measures and their predicted short-term evolution are compared for all alternative control strategies (e.g. travel routes or traffic signal timings) according to control objectives. On this basis, a strategy that leads to an optimal performance of the traffic control is selected for implementation.

An important requisite for the on-line simulation is a suitable model that reproduces the evolution of road traffic. Such model should present a well-balanced trade-off between accuracy and computational complexity to enable the on-line processing of traffic data. A representation of uncertainty is necessary in the model to take into account the inherent complexity of traffic processes as well as imprecise character of traffic measurements. Furthermore, the individual features of vehicles have to be modelled as they have a significant influence on the performance of traffic control. The traffic model also has to provide interfaces for many different sources of traffic data that have become available recently (e.g. vision-based monitoring systems [11] and vehicular sensor networks [7]). To facilitate the on-line simulation, the traffic model has to be adjusted in order to maintain consistency between simulated and measured traffic.

The issues discussed above have motivated the development of a fuzzy cellular traffic model, which was formulated as a hybrid system combining cellular automata and fuzzy calculus [10]. The fuzzy cellular model is based on a cellular automata approach to traffic modelling that ensures the accurate simulation of real traffic phenomena (for an overview see [8]). The original feature distinguishing this model from the other cellular models is that vehicle position, its velocity and other parameters are modelled by fuzzy numbers. Moreover, the rule of model transition from one time step to the next is also based on fuzzy definitions of basic arithmetic operations. The application of fuzzy calculus helps to deal with imprecise traffic data and describe uncertainty of the simulation results. These facts along with low computational complexity make the model suitable for the on-line processing of traffic data.

Hybrid systems that combine the cellular automata and fuzzy logic are typically referred to as fuzzy cellular automata (FCA) [2]. FCA-based models have found many applications in the field of complex systems simulation e.g. [1], [12]. A road traffic model of this kind has been proposed in [3]. In such models, the local update rule of classical cellular automata is usually replaced by a fuzzy logic system consisting of fuzzy rules, fuzzification, inference, and defuzzification mechanisms. A different approach is used in this paper: current states of the cells are determined by fuzzy sets and calculus with fuzzy numbers is involved in the update operation. The innovative features of the proposed methodology are the elimination of information loss caused by defuzzification and the incorporation of uncertainty in simulation results.

In this paper the fuzzy cellular model is applied to on-line simulation in order to evaluate the performance of road traffic control. To reduce the computational effort associated with on-line simulation, the fuzzy cellular model is implemented using a concept of ordered fuzzy numbers [4]. Algebra of the ordered fuzzy numbers is significantly more efficient than the solution based on classical fuzzy numbers and extension principle applied in [10]. These advantages enable computationally efficient evaluation of performance measures that are represented by means of fuzzy numbers. As it is shown in this paper, this representation is convenient for the determination of uncertainty in traffic control decisions.

The introduced method provides significant improvement when comparing with the state-of-the-art techniques. Existing methods use traffic parameters that describe queues or groups of vehicles rather than individual cars (e.g. flow volume, density, and mean velocity) [6], [14]. However, modern sensing platforms offer traffic data concerning the parameters of particular vehicles (position, velocity, acceleration, class, etc.) [7]. These data cannot be fully utilised when using the existing methods. In the proposed approach the precision of traffic state description is variable and can be adjusted to match the precision of available traffic data.

The rest of the paper is organised as follows: Section 2 describes the implementation of fuzzy cellular traffic model. Basic measures for the evaluation of traffic control performance are defined in Section 3. In Section 4, an issue of imprecise traffic data processing is discussed. Section 5 contains the results of an experimental study. Finally, conclusions are given in Section 6.

## 2 Traffic model for on-line data processing

A traffic lane in the fuzzy cellular model is divided into cells that correspond to the road segments of equal length. The traffic state is described in discrete time steps. These two basic assumptions are consistent with those of the Nagel-Schreckenberg cellular automata model. Thus, the calibration methods proposed in [9] are also applicable here for the determination of cells length and vehicles properties. A novel feature in this approach is that vehicle parameters are modelled using ordered fuzzy numbers [4]. The rule of model transition from one time step to the next is also based on fuzzy definitions of basic arithmetic operations.

Let us make an assumption that only triangular and trapezoidal fuzzy sets will be used in the traffic model. Hereinafter, all the ordered fuzzy numbers are represented by four integers and the following notation is used: $A = (a^{(1)}, a^{(2)}, a^{(3)}, a^{(4)})$. This notation is suitable for both triangular as well as trapezoidal membership functions. The arithmetic operations of addition and subtraction as well as the minimum function are computed for the ordered fuzzy numbers using the following definition:

$$o(A,B) = (o(a^{(1)},b^{(1)}), o(a^{(2)},b^{(2)}), o(a^{(3)},b^{(3)}), o(a^{(4)},b^{(4)})), \qquad (1)$$

where $A$, $B$ are the ordered fuzzy numbers and $o$ stands for an arbitrary binary operation.

The road traffic stream is represented in the fuzzy cellular model as a set of vehicles. A vehicle $n$ is described by its position $X_{n,t}$, velocity $V_{n,t}$ (in cells per time step), maximal velocity $V_{\max,n}$ and acceleration $A_n$. All these quantities are expressed by fuzzy numbers. The position $X_{n,t}$ is a fuzzy number defined on the set of cells indexes. Velocity of vehicle $n$ at time step $t$ is computed as follows:

$$V_{n,t} = \min\{V_{n,t-1} + A_n(V_{n,t-1}), G_{n,t}, V_{\max,n}\}, \quad G_{n,t} = X_{n-1,t} - X_{n,t} - (1,1,1,1), \qquad (2)$$

where $G_{n,t}$ is the fuzzy number of free cells in front of a vehicle $n$, $n - 1$ denotes the number corresponding to the lead vehicle and $n$ that of the following vehicle. If there is no lead vehicle in front of the vehicle $n$ then $G_{n,t}$ is assumed to be equal to $V_{\max,n}$. Acceleration is defined as a function of velocity to enable implementation of a slow-to-stop rule that exhibits more realistic driver behaviour [8]. After determination of velocities for all vehicles, their positions are updated as follows:

$$X_{n,t+1} = X_{n,t} + V_{n,t}. \qquad (3)$$

The preceding formulation of the fuzzy cellular model is illustrated in Fig. 1, which shows the results of numerical motion simulation of two accelerating vehicles during three time steps. Fig. 1 presents membership functions of the fuzzy numbers representing vehicles positions, the remaining parameters of vehicles are listed in Table 1. The maximal velocity of vehicles in this example is set as follows: $V_{max,1} = V_{max,2} = (1,2,2,3)$. The acceleration for both vehicles can take two fuzzy values depending on the vehicle's velocity: $A_n(V_{n,t-1}) = (1,1,1,1)$ if $V_{n,t-1} = V_{max,n}$ or $V_{n,t-1} = V_{max,n} - (1,0,0,0)$ and $A_n(V_{n,t-1}) = (0,1,1,1)$ else.

Since the fuzzy cellular model is designed to be used for the performance evaluation of traffic control, it has to take into account the status of traffic control operations. In case of a traffic signal control, drivers' reactions to traffic signals need to be considered. The influence of traffic signalisation is simulated in this study by introducing phantom vehicles in cells corresponding with the locations of traffic signals. Actual as well as maximal velocity of a phantom vehicle is always equal zero (0,0,0,0). Such vehicle is inserted into simulation during the red signal period and removed when the green signal is active.

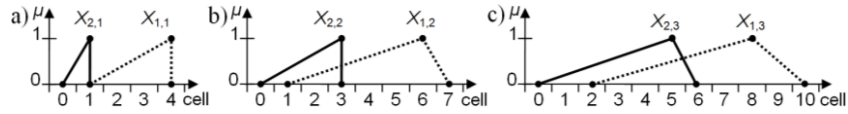

**Fig. 1.** Motion simulation of two vehicles: a) $t = 1$, b) $t = 2$ c) $t = 3$

**Table 1.** Vehicles parameters for the simulation illustrated in Fig. 1.

| $t$ | $X_{1,t}$ | $A_{1,t}$ | $G_{1,t}$ | $V_{1,t}$ | $X_{2,t}$ | $A_{2,t}$ | $G_{2,t}$ | $V_{2,t}$ |
|---|---|---|---|---|---|---|---|---|
| 0 | (1,2,2,2) | (0,1,1,1) | $V_{max,1}$ | (0,2,2,2) | (0,0,0,0) | (0,1,1,1) | (0,1,1,1) | (0,1,1,1) |
| 1 | (1,4,4,4) | (0,1,1,1) | $V_{max,1}$ | (0,2,2,3) | (0,1,1,1) | (0,1,1,1) | (0,2,2,2) | (0,2,2,2) |
| 2 | (1,6,6,7) | (1,1,1,1) | $V_{max,1}$ | (1,2,2,3) | (0,3,3,3) | (0,1,1,1) | (0,2,2,3) | (0,2,2,3) |
| 3 | (2,8,8,10) | (1,1,1,1) | $V_{max,1}$ | (1,2,2,3) | (0,5,5,6) | (1,1,1,1) | (1,3,3,3) | (1,2,2,3) |

## 3   Performance measures

There are several different measures available that can be employed for the evaluation of traffic control performance e.g.: average delay per vehicle, maximum individual delay, percentage of cars that are stopped, average number of stops, queue length, throughput of intersections, and travel time. In this section an algorithm is provided for computing the basic performance measures in the fuzzy cellular model. For the formal presentation of the algorithm a function $S_C$ is defined that acts on directed fuzzy numbers:

$$S_C(A) = (s^{(1)}, s^{(2)}, s^{(3)}, s^{(4)}), \quad s^{(i)} = \begin{cases} 0, & |A_c| < 5-i \\ 1, & |A_c| \geq 5-i \end{cases}, \quad (4)$$

where $A_C$ is a set of integers used in the notation of fuzzy number $A$, which satisfy the condition denoted by $C$: $A_C = \{a^{(i)} \in \{a^{(1)}, a^{(2)}, a^{(3)}, a^{(4)}\} : a^{(i)} \text{ satisfies } C\}$.

Function $S_C(A)$ allows us to determine a level of confidence that the condition $C$ is satisfied by $A$. The value of this function is (0, 0, 0, 0) if the condition $C$ is false for $A$ and (1, 1, 1, 1) if the condition is true. Any other combination of $s^{(i)}$ means that the condition is partially satisfied by $A$. For example, let us check the condition "vehicle is stopped" for the second time step of the simulation presented in Fig. 1 ($t = 2$). Such a condition can be written as $V_{n,2} = 0$ and the corresponding form of the function (4) is $S_{=0}(V_{n,2})$. The values of this function are as follows: $S_{=0}(V_{1,2}) = (0, 0, 0, 0)$ for the lead vehicle and $S_{=0}(V_{2,2}) = (0, 0, 0, 1)$ for the following vehicle (see Table 1). It means that the lead vehicle is not stopped and there is a possibility that the following vehicle is stopped.

Traffic performance measures can be now formulated in terms of the fuzzy cellular model, using the function $S_C$. The most commonly used criteria of performance are average delay and average number of stops. Average delay in time steps per vehicle is defined as:

$$D = \frac{1}{N} \sum_n \sum_t S_{=0}(V_{n,t}), \qquad (5)$$

where $N$ is the total number of vehicles. The quotient $A/N$ of an ordered fuzzy number $A$ and an integer $N$ is computed according to formula: $A/N = (a^{(1)}/N, a^{(2)}/N, a^{(3)}/N, a^{(4)}/N)$. Additionally, the values of $a^{(i)}/N$ are rounded to the nearest integers.

A stop of a vehicle is the situation in which the velocity of a vehicle is higher than zero for the previous time step of simulation ($t - 1$) and it is zero for the current time step ($t$). Thus, the average number of stops is given by:

$$S = \frac{1}{N} \sum_n \sum_t \min\{S_{>0}(V_{n,t-1}), S_{=0}(V_{n,t})\}. \qquad (6)$$

If a vehicle $n$ is stopped in queue, then the number $G_{n,t}$ of free cells in front of it is zero. In this study the length of a vehicle is assumed to be equal one cell, therefore the average queue length in cells can be computed using the following formula:

$$Q = \frac{1}{T} \sum_n \sum_t S_{=0}(G_{n,t}), \qquad (7)$$

where $T$ is the number of steps in the analysed time period of traffic simulation.

The main advantage of the presented model relies on the fact that the performance estimation of traffic control is computationally efficient and the uncertainty of the results is taken into account. The results concerning traffic performance are represented by means of fuzzy numbers. As it is shown in Section 5, this representation is convenient for the determination of uncertainty in control decisions.

## 4 Modelling of an imprecise traffic information

Precise data on the parameters of each particular vehicle are often hardly available due to lack of sensing devices and high transmission costs. This section discusses an issue of imprecise traffic data processing with application of the fuzzy cellular model.

Let us assume that the data on vehicles locations are delivered with a precision, which is insufficient to place vehicles in specific cells. It means that the available traffic information describes the number of vehicles that are located in a given segment of cells. In Fig. 2 an example is illustrated, which corresponds to information indicating that there are three vehicles present in the segment of fifteen cells. This information is introduced in the model through assumption of the four instances that are shown in Fig. 2 a-d. For the instances a and b it was assumed that the vehicles are equally spaced and are moving with maximal velocity. The instances c and d represent the least possible situations in which the vehicles are stopped in a queue. A fusion of these four instances is achieved by using ordered fuzzy numbers to describe the vehicles positions (Fig. 2 e-g).

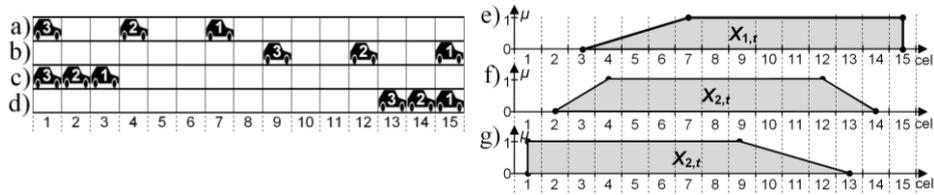

**Fig. 2.** Representation of imprecise traffic information in fuzzy cellular model

To generalize the example discussed above, let us consider $N$ vehicles located in a road segment that consists of cells $c_S, ..., c_E$. A position of vehicle $n$ ($n = 1, ..., N$) is defined by the following equations:

$$x_{n,t}^{(1)} = c_S + N - n, \quad x_{n,t}^{(2)} = c_S + \sum_{i=n+1}^{N} \min\left\{\lfloor L/N \rfloor, v_{\max,i}^{(2)} + 1\right\}, \quad (8)$$
$$x_{n,t}^{(3)} = c_E - \sum_{i=2}^{n} \min\left\{\lfloor L/N \rfloor, v_{\max,i}^{(3)} + 1\right\}, \quad x_{n,t}^{(4)} = c_E + 1 - n,$$

where $L = c_S - c_E + 1$ is the number of cells in the road segment (precision unit).

The above definition allows the imprecise data to be utilised for the evaluation of traffic control performance. It demonstrates also the applicability of the introduced approach for modelling the imprecise traffic information.

## 5 Case study

The fuzzy cellular model was applied to performance evaluation of traffic control at a road work zone. Fig. 3 shows a schematic layout of the simulated situation. Average delay was analysed in this experiment for two different control strategies. At the beginning of each simulation vehicles were randomly placed on road lanes approaching the work site (lanes *A* and *B*). In the first strategy a green signal is displayed for the traffic lane *A*. Afterwards, when all vehicles vacate this lane, the green light is activated for the opposite direction (lane *B*). The second control strategy assumes that an inverse traffic signal sequence is used (the green signal for lane *B* is activated at first). For both strategies the average delay per vehicle was calculated as a fuzzy number according to equation (5). Simulations were executed for various numbers of vehicles generated in the traffic lanes and for different precision units. The precision unit *L* was defined as the number of cells in segments that are used for determination of the initial vehicles locations (see Section 4).

The average delays $D_i = (d_i^{(1)}, d_i^{(2)}, d_i^{(3)}, d_i^{(4)})$ evaluated for both control strategies ($i$ = 1, 2) have to be compared in order to select the optimal strategy for implementation. An example of such comparison is presented in Fig. 4 (a and b). The simulation was executed for 50 and 45 vehicles placed in lanes *A* and *B* respectively ($N_A$=50, $N_B$=45). It can be observed that the optimal control strategy cannot be selected without ambiguity and the uncertainty of this selection arises when higher precision units are in use.

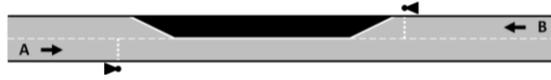

**Fig. 3.** Lane closure on two-lane road using traffic signals

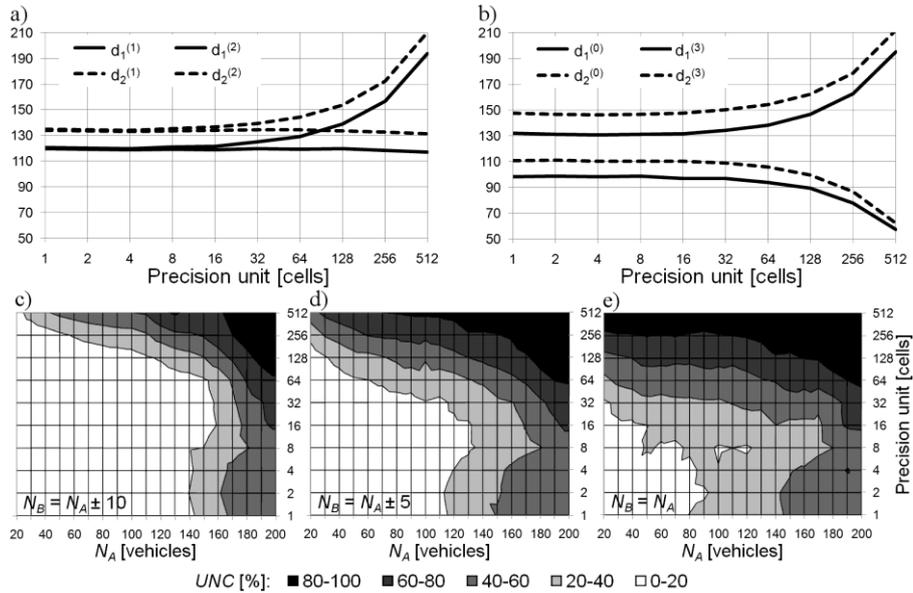

**Fig. 4.** Experimental results: a) and b) average delays, c) – d) uncertainty of strategy selection

The uncertainty of strategy selection was determined using the following formula:
$$UNC = 1 - \max\{P(D_1 < D_2), P(D_2 < D_1)\} + \min\{P(D_1 < D_2), P(D_2 < D_1)\}, \quad (9)$$

where *P(C)* denotes a probability of satisfying condition *C*. The probabilities in equation (9) were computed using a method of fuzzy numbers comparison proposed in [9]. The uncertainty takes a value between 0 and 1 (0 – 100 [%]). Contour plots in

Fig. 4 present the results of uncertainty determination that were obtained assuming various absolute differences between the numbers of vehicles placed in traffic lanes $A$ and $B$ ($|N_A - N_B|$). The differences are equal 10, 5 and 0 for Fig. 4 c, d and e respectively. It can be seen that the uncertainty increases when the difference $|N_A - N_B|$ is decreased. The uncertainty level is similar for precision units in a range from 1 to 16 cells; however, further decrease in the traffic data precision causes a significant increase of the uncertainty. This way of analysis allows us to determine the minimal precision of traffic data, which is necessary to select the optimal control strategy using the proposed performance evaluation method.

## 6 Conclusions

The fuzzy cellular model enables the evaluation of performance measures for road traffic control. Imprecise traffic data can be utilised in the proposed hybrid modelling approach. The results of performance evaluation are represented in terms of ordered fuzzy numbers. This representation is suitable for the uncertainty determination in traffic control decisions that are based on the performance comparison for several candidate control strategies. The uncertainty level can be used to decide if the currently available traffic data are sufficient for the selection of optimal control strategy. It allows the precision level of traffic measurements to be dynamically adapted to both the current traffic situation and the requirements of traffic control system.